\documentclass{article}
\usepackage{mypreprint}
\usepackage{lipsum}

\begin{document}

\twocolumn[
\mytitle{DDDM: a Brain-Inspired Framework for Robust Classification}


\mysetsymbol{equal}{*}

\begin{myauthorlist}
\myauthor{XiYuan Chen}{equal,ion,ucas}
\myauthor{Xingyu Li}{equal,bsbii}
\myauthor{Yi Zhou}{ustc,bsbii}
\myauthor{Tianming Yang}{ion,bsbii}
\end{myauthorlist}

\myaffiliation{ion}{Institute of Neuroscience, Center for Excellence in Brain Science and Intelligence Technology, Chinese Academy of Sciences, Shanghai, China}
\myaffiliation{ucas}{University of Chinese Academy of Sciences, Beijing, China}
\myaffiliation{bsbii}{Shanghai Center for Brain Science and Brain-Inspired Technology, Shanghai, China}
\myaffiliation{ustc}{National Engineering Laboratory for Brain-inspired Intelligence Technology and Application, School of Information Science and Technology, University of Science and Technology of China, Hefei, China}

\mycorrespondingauthor{Yi Zhou}{yizhoujoey@gmail.com}
\mycorrespondingauthor{Tianming Yang}{tyang@ion.ac.cn}

\mykeywords{Brain-Inspired Intelligence, DDM, Robustness}

\vskip 0.3in
]



\printAffiliationsAndNotice{\myEqualContribution} 

\begin{abstract}
Despite their outstanding performance in a broad spectrum of real-world tasks, deep artificial neural networks are sensitive to input noises, particularly adversarial perturbations. On the contrary, human and animal brains are much less vulnerable. In contrast to the one-shot inference performed by most deep neural networks, the brain often solves decision-making with an evidence accumulation mechanism that may trade time for accuracy when facing noisy inputs. The mechanism is well described by the Drift-Diffusion Model (DDM). In the DDM, decision-making is modeled as a process in which noisy evidence is accumulated toward a threshold. Drawing inspiration from the DDM, we propose the Dropout-based Drift-Diffusion Model (DDDM) that combines test-phase dropout and the DDM for improving the robustness for arbitrary neural networks. The dropouts create temporally uncorrelated noises in the network that counter perturbations, while the evidence accumulation mechanism guarantees a reasonable decision accuracy. Neural networks enhanced with the DDDM tested in image, speech, and text classification tasks all significantly outperform their native counterparts, demonstrating the DDDM as a task-agnostic defense against adversarial attacks.
\end{abstract}

\section{Introduction}

\begin{figure*}[!ht]
    \centering
    \includegraphics[width=0.88\textwidth]{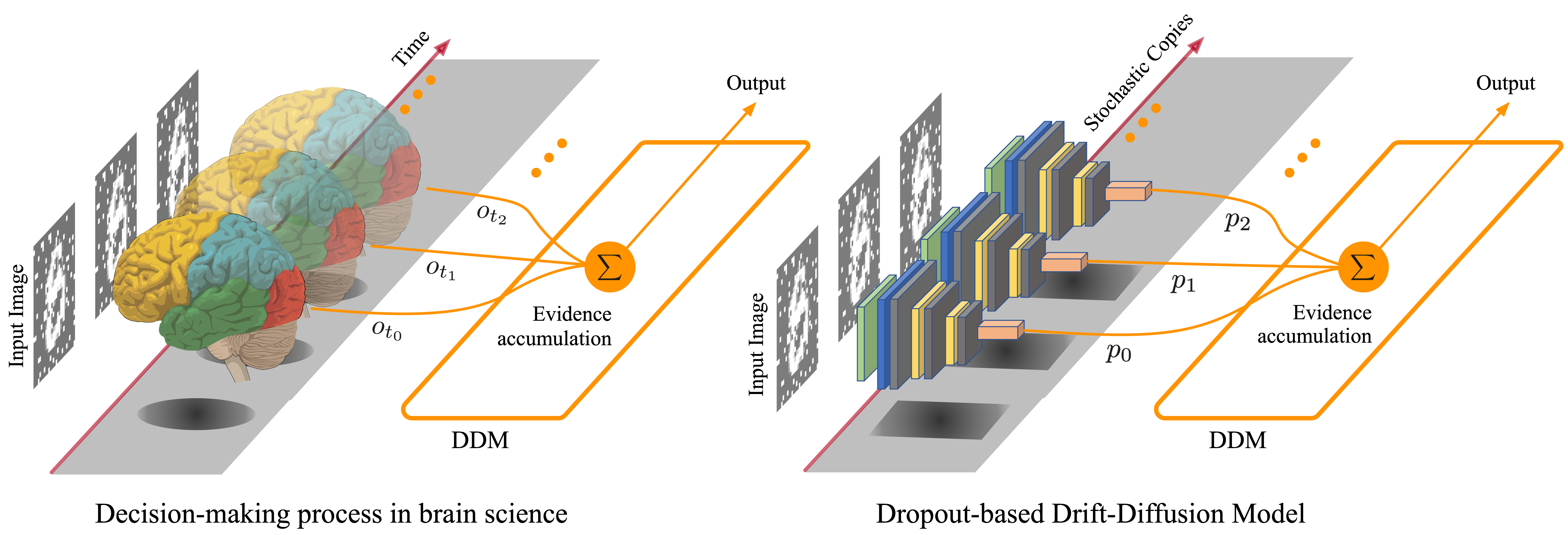} 
    \caption{ 
    Schematic diagram of the Dropout-based Drift-Diffusion Model (right) and its brain science counterpart (left). In DDDM, outputs $p_i$ from the stochastic copies of an arbitrary neural network simulate the noisy temporal neural signal $o_{t_i}$ in the brain. In both, the series of outputs/signals are passed to a similar evidence accumulation mechanism (as described by the DDM) for a robust output.}
    \label{fig1}
\end{figure*}

Deep learning has shown outstanding performance in many areas, including image, audio
, and text classification. Nevertheless, deep learning lacks robustness; that is, deep artificial neural networks are susceptible to manipulations of inputs. Small perturbations on inputs that are imperceptible to humans may result in significant differences in outputs of deep neural networks~\cite{goodfellowExplainingHarnessingAdversarial2015a}. Such susceptibility is not only of theoretical interest but also represents a severe practical problem, which causes security issues in real-world applications, including face recognition and autonomous driving.

Recently, a huge amount of effort has been dedicated to addressing this issue. 
Adversarial training~\cite{madryDeepLearningModels2019} is considered to be the most promising. It adds adversarial samples to the training set and automatically lets the network model adapt to them through learning. Despite these efforts, current defenses against adversarial attacks are still far from satisfactory. Most defenses only target one or a few types of attacks and are often restricted to particular tasks, such as image classification. Consequently, it is unclear whether they can be applied to other kinds of attacks or tasks. 

In contrast, humans and animals are much less vulnerable. Even though 
time-limited human subjects can be misled by deliberately crafted images,
they successfully restore accuracy by spending more time facing these uncertain and noisy inputs~\cite{elsayedAdversarialExamplesThat2018a}. The more uncertain the input is, the longer the brain takes to make a decision. This ability of trading decision speed for accuracy has been well studied in humans and animals by researchers in neuroscience and psychology~\cite{ratcliffModelingResponseTimes1998,roitmanResponseNeuronsLateral2002}. In particular, the celebrated Drift-Diffusion Model (DDM) and its variants stand as one of the most successful models that account for the decision-making process in humans and animals. The DDM describes a sequential sampling process in which signals with both noises from the external sources and the internal neuronal noises are accumulated as evidence for decisions. The decision is made once the accumulated evidence reaches a pre-determined threshold. A high threshold leads to accurate but slow decisions, while a low threshold results in fast but inaccurate decisions. Neuronal activities in decision-making-related brain areas exhibit response patterns described by the DDM~\cite{roitmanResponseNeuronsLateral2002}.

\footnotetext[1]{
    Implementation: https://github.com/XiYuan68/DDDM.
}

The ability to trade speed for accuracy characterizes one of the most striking differences between biological and artificial neural networks. Enhancing artificial neural networks with a dynamic inference process similar to those in human and animal brains would potentially improve their robustness. 
Guided by this intuition,
we propose the Dropout-based Drift-Diffusion Model (DDDM) as a general framework for enhancing the robustness of arbitrary neural networks. Depicted in Figure 1, in contrast to adversarial training, which adds noises to inputs during the training phase, DDDM adds noises to neural networks via random dropouts during the test phase, resulting in multiple stochastic copies of the original one. Then, DDDM employs a DDM-like evidence accumulation mechanism to make the decision based on these outputs.

We conduct experiments on three types of datasets, including MNIST~\cite{lecunGradientbasedLearningApplied1998} and CIFAR10~\cite{krizhevsky_learning_2009} for image classification, the SpeechCommands~\cite{wardenSpeechCommandsDataset2018} for audio classification, and the IMDB dataset~\cite{maasLearningWordVectors2011} for text classification. We compare the networks' performance with and without the DDDM against a variety of adversarial attacks, including both the white-box and the black-box ones. The experimental results show that the DDDM improves the robustness of the network, providing defense against all adversarial attacks tested on all three datasets. 
\section{Related Work}

Adversarial training (AT)~\cite{goodfellowExplainingHarnessingAdversarial2015a,madryDeepLearningModels2019} is one of the most popular adversarial defense methods. 
It augments the clean data with adversarial ones during the trianing process.
There are works that try to improve robustness by introducing additional noise in their models. For example, \cite{guoCounteringAdversarialImages2018} applied transformation on images before feeding the network, including image quilting, total variance minimization, JPEG compression, and bit-depth reduction. Similar to our work, \cite{dhillonStochasticActivationPruning2018} randomly dropped neurons with a weighted distribution. Note that all the defense methods mentioned above focus on 
either filtering out adversarial perturbations or 
creating obfuscated gradients~\cite{athalyeObfuscatedGradientsGive2018}, leaving room for more sophisticated attacks.
On the contrary, our framework injects randomness into the network model during the testing phase, and the resulting stochastic predictions are agnostic to the type of adversarial attacks. Furthermore, a temporal evidence accumulation mechanism is employed to offset the accuracy reduction due to the added randomness and retain a good performance with robustness.

In neuroscience, recent advances in the study of perceptual decision-making have started to reveal the underlying neural mechanism~\cite{goldNeuralBasisDecision2007}. 
The random-dot motion (RDM) direction discrimination task is one of the most popular tasks for studying decision-making with noisy evidence~\cite{roitmanResponseNeuronsLateral2002}.
In this task, subjects are asked to judge whether a patch of noisy dots moves toward left or right. The subjects' decision accuracies and response times can be well described by the DDM~\cite{ratcliffDiffusionDecisionModel2008}. 
Furthermore, it was demonstrated that neurons in the posterior parietal cortex increased their firing when evidence toward their favored choice was being accumulated~\cite{roitmanResponseNeuronsLateral2002}. 
The response increase correlated with the quality of evidence, which may be quantified as the log-likelihood ratio~\cite{goldNeuralBasisDecision2007,yangProbabilisticReasoningNeurons2007}, and reflected the evidence accumulation process. 
Additional studies extended DDM into probabilistic inferences based on images and identified a prefrontal-parietal circuitry that carries out evidence accumulation process in probabilistic inferences~\cite{yangProbabilisticReasoningNeurons2007}. 
Response patterns that reflect evidence accumulation were also found in many other brain regions, including multiple regions in the prefrontal cortex and the basal ganglia, suggesting evidence accumulation is a universal decision-making mechanism employed by the brain~\cite{goldNeuralBasisDecision2007}.
DDM and the related sequential sampling models in neuroscience describe the brain's solutions for optimizing decision-making with noisy inputs. However, the DDM has not been used together with deep neural networks in the machine learning field and has not been considered as a defense mechanism against adversarial attacks.
\section{Model Description}

\subsection{Test-Phase Dropout}
Dropout was initially proposed by~\cite{hintonImprovingNeuralNetworks2012} to address the problem of overfitting. In this paper, we use it to introduce randomness in the system. It simulates the internal neuronal noise in 
synaptic transmissions~\cite{rusakov_noisy_2020}.
With dropouts applied temporally, the original network model effectively extends to an ensemble of stochastic copies, whose outputs serves as the evidence. Let $h_{a,b}$ be an arbitrary deep neural network with one or several dropout layers. The dropout layers share the same dropout rate. Subscript $a$ denotes the dropout rate used for model training and validation, and $b$ denotes the dropout rate at the test phase. We refer to $h_{a,b}$ as the dropout classifier in the following context to emphasize the involvement of the test-phase dropout mechanism.

\subsection{Drift-Diffusion Model}

\begin{figure}[!t]
    \centering
    \includegraphics[width=0.8\columnwidth]{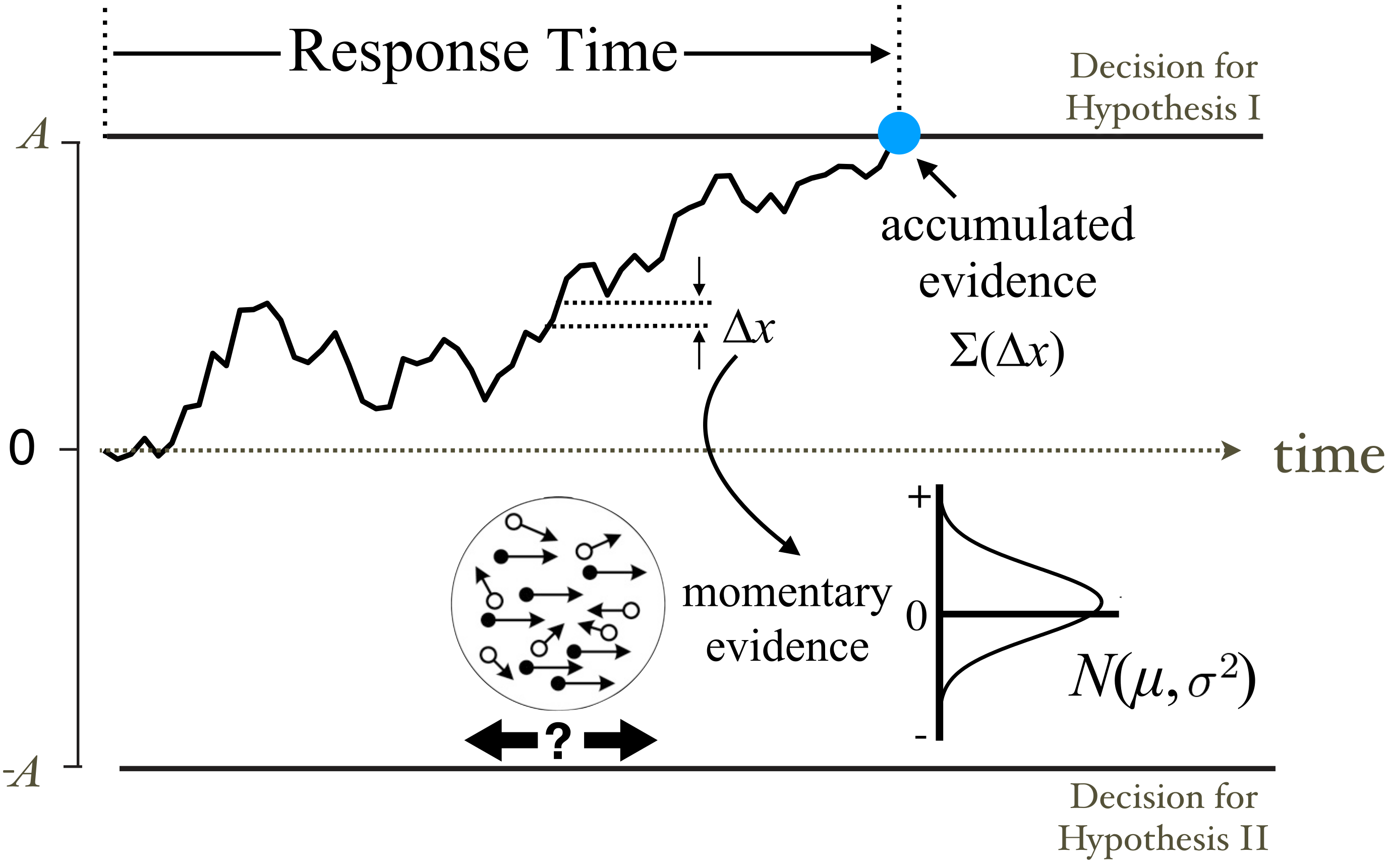} 
    \caption{
    A diagram of the drift-diffusion model (DDM).
    At each moment, 
    the noisy momentary evidence is accumulated until it hits one of the two decision thresholds ($A$ and $-A$). The decision threshold and the mean and the variance of the momentary evidence together determine the speed and accuracy of the decision.}
    \label{DDM}
\end{figure}

In a typical two-choice scenario, the canonical Drift-Diffusion Model (DDM) (see~\Cref{DDM}) describes decision-making as a process in which evidence that is noisy and fluctuates over time (momentary evidence) is accumulated toward a preset threshold $A$. Mathematically, DDM corresponds to the following stochastic differential equation
\begin{equation}\label{ddm2c}
    \tau \dot{X} = \mu + \xi(t),
\end{equation}
where $X$ is known as the decision variable representing the cumulated evidence, and $\mu$ is the drift or the mean of the noisy momentary evidence. The Gaussian process $\xi(t)$, with mean $0$ and variance $\sigma^2$, captures the uncertainty in the evidence. Coefficient $\tau$ is the characteristic time scale. The decision variable $X$ evolves until reaching one of the two thresholds $\pm A$ at which the corresponding decision is made.
These thresholds correspond to the choice options, and the time needed for the accumulated evidence to reach a particular threshold reflects the response time (RT). 

The canonical DDM can be extended to decisions with more than two alternatives.
\Cref{ddm2c} can be derived from a set of stochastic differential equations~\cite{roxin_driftdiffusion_2019}.
Following this route, it is straightforward to increase the number of choices by extending the number of differential equations. Another route to extend the DDM to the multi-choice domain models the decision-making as a Bayesian multiple sequential probability ratio test (MSPRT).
It has been demonstrated the two routes are equivalent~\cite{roxin_driftdiffusion_2019}.

For our purpose to integrate the DDM with the test-phase dropout classifier, we use the Bayesian MSPRT route. Suppose there are $n$ alternatives, and $z_i(t)$ denotes the instantaneous evidence for choice $i$. $z_i(t)$ is drawn from a Gaussian distribution with mean $M_i$ and variance $\sigma^2$. Let $\mathcal{O}(T):=\{z_i(t): 1\le i \le n, 1\le t \le T\}$ be the collection of observations up to a time $T$, and $H_i$ denote the hypothesis that choice $i$ has the largest mean. Within the Bayesian MSPRT framework, one seeks to make a decision based on the posterior $\text{Pr}(H_i|\mathcal{O}(T))$. With defining
\begin{equation*}
    X_k(t) := \sum_i^k y_i(t) - k y_{k+1}(t), \quad 1\le k \le n-1,
\end{equation*}
where $y_i(t):=\int_0^t dt'\,z_i(t')$
and $X_k$ is the decision variable for the $k$th class, measuring how likely the original input belongs to class k given the accumulated evidence.
It can be shown that $X_k$ satisfies a stochastic differential equation of the same type as~\Cref{ddm2c}.
We refer the interested readers to~\cite{roxin_driftdiffusion_2019} for a complete deduction.

\begin{table*}[!ht]
    \centering
    \setlength{\tabcolsep}{1.4 mm}
    \begin{tabular}{cccccccccc}
        \toprule
                           & clean       & FGSM        & PGD         & $L_2$ C\&W    & $L_2$ DF         & Salt\&Pepper & Uniform     & Spatial     & Square      \\\midrule
        $h_{0,0}$ (\%)     & 98.10       & 12.10       & 0.00        & 1.70          & 42.70            & 0.20         & 94.50       & 0.40        & 19.50       \\\midrule
        $h^*_{a,b}$ (\%)   & 99.20       & 50.70       & 43.30       & 96.70         & 67.90            & 96,90        & 97.10       & 34.80       & 78.00       \\
        {\small $(a, b)^*$}         & {\small $(0.2,0.0)$} & {\small $(0.0,0.6)$} & {\small $(0.0,0.6)$} & {\small $(0.4,0.4)$}   & {\small $(0.2,0.6)$}      & {\small $(0.2,0.4)$}  & {\small $(0.2,0.2)$} & {\small $(0.6,0.8)$} & {\small $(0.0,0.2)$} \\\midrule
        $H^*_{a,b}$ (\%)   & {99.16}     & {73.12}     & {74.85}     & {98.89}       & {95.04}          & {98.89}      & {98.24}     & {65.22}     & {88.93}     \\
        {\small $(a, b)^*$}         & {\small $(0.4,0.2)$} & {\small $(0.0,0.6)$} & {\small $(0.0,0.8)$} & {\small $(0.2,0.4)$}   & {\small $(0.6,0.8)$}      & {\small $(0.4,0.4)$}  & {\small $(0.2,0.4)$} & {\small $(0.4,0.8)$} & {\small $(0.0,0.6)$} \\
        \midrule\midrule
        $h_{a^*,b^*}$ (\%) & 85.90       & 49.70       & 36.40       & 85.40         &  67.90           &  86.70       & 79.10       &  30.20      &  64.10      \\\midrule
        $H_{a^*,b^*}$ (\%) & 98.20       & 71.09       & 51.68       & 97.94         &  92.30           &  98.08       & 97.36       &  36.73      &  87.15      \\
        \bottomrule
    \end{tabular}
    \caption{\label{mnist}Performance of the DDDM under eight adversarial attacks on MNIST. 
    The optimal performance of the DDDM $H^*_{a, b}$ for individual attacks are compared to the ones of the dropout classifier $h^*_{a,b}$.
    The accuracy of the undefended model $h_{0,0}$ is included as the baseline. 
    For each case, we show the prediction accuracy of the individual optimal models $h^*_{a,b}$/$H^*_{a, b}$ as well as their corresponding optimal dropout rate pair $(a,b)^*$.
    We also include the overall performance of the DDDM $H_{a^*, b^*}$ and the dropout classifier $h_{a^*,b^*}$.
    The global optimal dropout rate pair $(a^*,b^*)$ is chosen as $(0.2,0.6)$.
    }
\end{table*}

\subsection{Dropout-based Drift-Diffusion Model}

There are two components in the design of the Dropout-based Drift-Diffusion model (DDDM). The first component turns the static one-shot inference process into a dynamic, noisy inference process. The noisy predictions from the stochastic copies of neural networks simulate the series of temporal signals in the biological brain. Presently, we implement the dynamic inference by introducing the test-phase dropout, which will randomly disturb the loss landscape and potentially make adversarial attacks harder to succeed. This results in a dropout classifier whose outputs are subject to further processing.

For the second component, we take the noisy outputs from the test-phase dropout classifier as evidence and apply DDM with the Bayesian MSPRT implementation.
More precisely, for a dropout classifier $h_{a,b}$, let $C$ be the number of categories and $x$ be an input. We arrange $L$ predictions to form a trial 
\begin{equation*}
    T = \begin{bmatrix}
            p_{11} & \cdots & p_{1C} \\
            \vdots & \ddots & \vdots \\
            p_{L1} & \cdots & p_{LC}
        \end{bmatrix},
\end{equation*}
where $p_{ij}$ is the $i$-th predicted confidence of $x$ belonging to label $j$. Let $T_i$ denote the $i$-th row vector in matrix $T$, we compute the posterior at time step $t$
\begin{equation*}
    P(j,t) := \text{Pr}(H_j|\{T_1,\dots, T_t\}), \, 1\le j \le C, 1\le t \le L.
\end{equation*}
Let $RT\le L$ be the smallest time step at which there exists a label $winner$ satisfying $P({winner}, RT)\ge A$. We say that $x$ belongs to the label $winner$ with a response time $RT$. If the threshold is never surpassed during the whole trial, we will force a decision by assigning the label winner to be the one with the largest decision variable at time step L, i.e., $P({winner}, L) = \max(P(1, L), P(2,L),\dots,P(C,L))$. 
In practice, we apply multiple trials and 
an appropriate decision threshold $A$ is chosen to offset the noise introduced by dropouts and ensure overall accuracy.
To compute the posterior $P(i,j)$, we apply Bayes theorem and assume equal priors for all hypotheses $H_i$. 
The problem is then reduced to computing the likelihoods $\text{Pr}(\{T_1,\dots, T_t\}|H_i)$, which are approximately evaluated over the clean training data.
The detailed description of our implementation can be found in the supplementary material.

In the following context, we will denote the dropout classifier equipped with DDM as $H_{a,b}$ and refer to it as DDDM for simplicity.

\section{Experiments and Results}

The primary goal of the present experiments is to evaluate the robustness of our DDDM model. 
Our experiments cover three different modalities of data, i.e., image, audio, and text. Under all types of inputs, the DDDM is demonstrated to be effective in defending adversarial attacks.

For all the tasks, we run experiments over different combinations of dropout rates $(a,b)$, both taking values from the set $\{0,0.2,0.4,0.6,0.8\}$. Therefore, twenty-five classifiers are tested in each case. 
In implementing DDM, 
we draw $100$ predictions from each classifier on each generated adversarial example. Then, $10$ trials of the length $L=25$ are randomly sampled from those predictions. These trials are fed into the evidence accumulation mechanism with a decision threshold $A=0.99$ for the final prediction.

\subsection{Images Classification}

\subsubsection{MNIST Dataset}
In the experiments, we adopt the same network architecture as in~\cite{madryDeepLearningModels2019} with an additional dropout layer that follows each layer except the input/output ones.
 
To comprehensively evaluate the effectiveness of DDDM in defending adversarial attacks, we considered four white-box attacks and four black-box attacks as listed below: the Fast Gradient Sign Method (FGSM) attack~\cite{goodfellowExplainingHarnessingAdversarial2015a}, the Projected Gradient Descent (PGD) attack~\cite{madryDeepLearningModels2019}, the $L_2$ Carlini and Wagner ($L_2$  C\&W) attack~\cite{carliniEvaluatingRobustnessNeural2017}, the $L_2$ DeepFool attack~\cite{moosavi-dezfooliDeepFoolSimpleAccurate2016a}, the Salt and Pepper attack, the $L_\infty$ uniform noise attack~\cite{rauberFastDifferentiableClippingAware2020}, the Spatial attack~\cite{engstromExploringLandscapeSpatial2019a} and the Square attack~\cite{andriushchenkoSquareAttackQueryEfficient2020}. The details of attack settings can be found in the supplementary material.

In~\Cref{mnist}, we compare the optimal DDDMs ($H^*_{a,b}$) with the optimal dropout classifiers ($h^*_{a,b}$) on individual attacks and the clean data. The undefended model ($h_{0,0}$) is included as the baseline. 
To demonstrate the effectiveness of DDDM under the combination of all attacks, we also include the results for $H_{a^*,b^*}$ and $h_{a^*,b^*}$, i.e., the DDDM and the dropout classifier with a single optimal dropout rate pair $(a^*,b^*)$ across all the attacks.
$(a^*,b^*)$ is determined with the criterion that the corresponding $H_{a^*,b^*}$ should try its best to avoid sacrificing the clean accuracy and, at the same time, provide as much defense as possible across different attacks.
The results indicate that the test-phase dropouts effectively neutralize the adversarial perturbations against all attacks in our experiments, and the DDM retains the accuracy through evidence accumulation. Furthermore, $H_{a^*,b^*}$ outperforms $h_{a^*,b^*}$ as well as $h_{0,0}$ on all the cases.

\begin{figure}[!t]
    \centering
    \includegraphics[width=0.83\columnwidth]{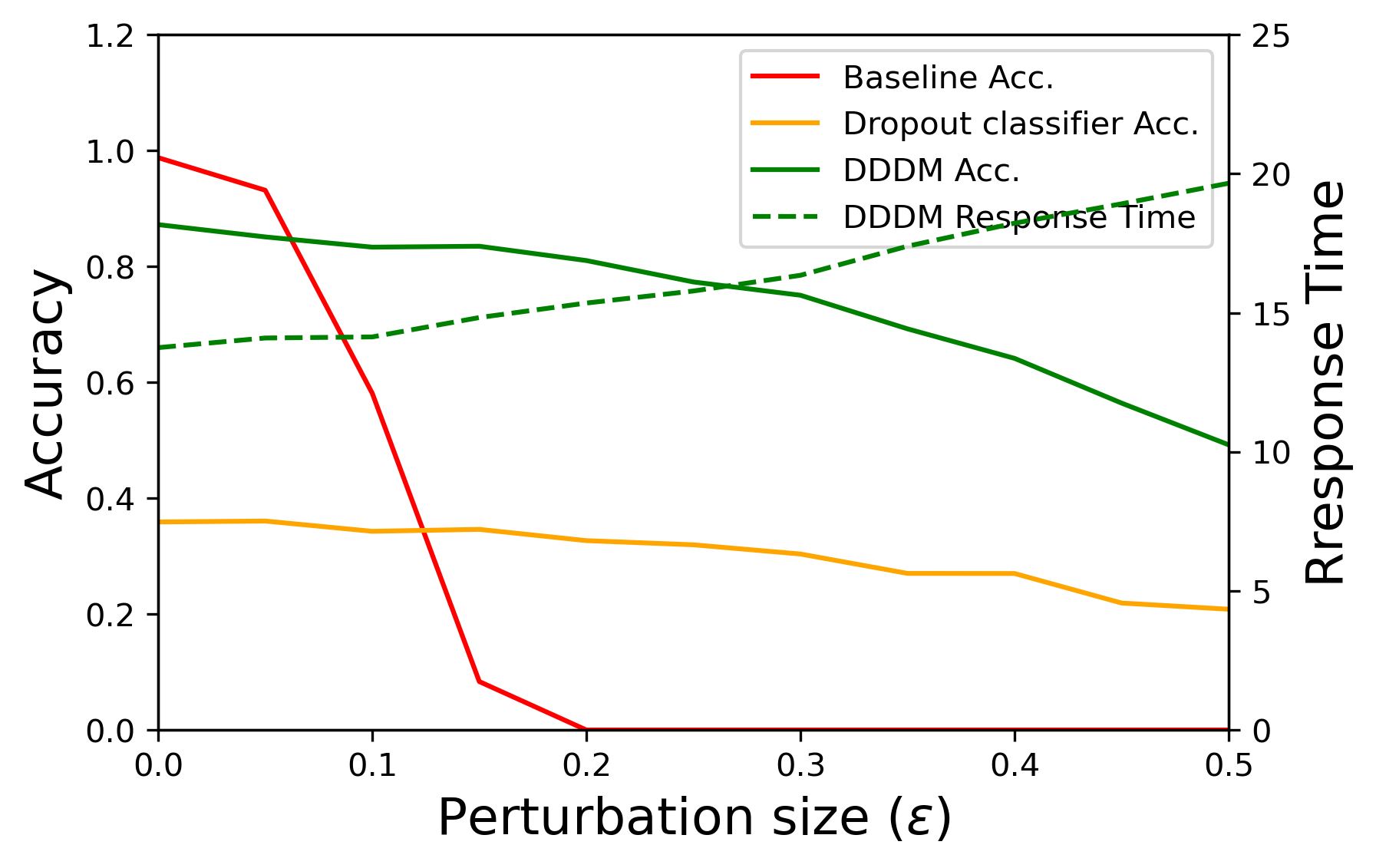} 
    \caption{Variation of accuracy (solid lines) and response time (the dashed line) against increasing perturbation size ($\epsilon$). 
    We compared three models: the baseline model $h_{0,0}$, the dropout classifier without DDM $h_{0.0, 0.8}$, and the dropout classifier with DDM $H_{0.0,0.8}$ under the PGD attack on MNIST.}
    \label{epsacc}
\end{figure}

\paragraph{Trade-off between Response Time and Accuracy.}
Animals and humans spend more effort in making difficult decisions, and the DDDM exhibits the same behavior. 
While the randomness in the test-phase dropout classifier neutralizes the adversarial perturbations, the DDM helps retain a decent accuracy by spending more time in inference.
To demonstrate this trade-off, we plot in~\Cref{epsacc} the accuracies and response times with respect to different adversarial perturbation sizes, comparing DDDM to the dropout classifier as well as the undefended baseline model.
The response time is measured by the number of forward passes.
The accuracy of the baseline model drops rapidly as the perturbation size increase. On the other hand, both the DDDM and the test-phase dropout classifier retain relatively high accuracy even when the perturbation is large. For small perturbations, the test-phase dropout classifier's accuracy curve lies well below the baseline model's curve. In contrast, the performance of DDDM dominates on almost the whole perturbation range except for near 0. Finally, the response time increases monotonically with perturbation level, demonstrating a trade for accuracy with time.

\begin{table}[!t]
    \centering
    \setlength{\tabcolsep}{1.4 mm}
    \begin{tabular}{ccccc}
        \toprule
                           & clean           & PGD             & $L_2$ DF            & Spatial          \\\midrule
        $h_{0,0}$ (\%)     & 89.71           & 5.24            & 67.96           & 8.93             \\\midrule
        $h^*_{a,b}$ (\%)   & 89.89           & 68.46           & 71.53           & 40.98            \\
        {\small $(a, b)^*$}         & {\small $(0.2,0.0)$}     & {\small $(0.8,0.8)$}     & {\small $(0.0,0.2)$}     & {\small $(0.4,0.8)$}      \\\midrule
        $H^*_{a,b}$ (\%)   & {89.67}         & {68.46}         & {79.01}         & {64.26}          \\
        {\small $(a, b)^*$}         & {\small $(0.6,0.4)$}     & {\small $(0.8,0.8)$}     & {\small $(0.0,0.8)$}     & {\small $(0.0,0.8)$}      \\
        \midrule\midrule
        $h_{a^*,b^*}$ (\%) & 86.88           & 35.87           & 67.57           & 36.92            \\\midrule
        $H_{a^*,b^*}$ (\%) & 89.30           & 36.76           & 70.00           & 37.59            \\
        \bottomrule
    \end{tabular}
    \caption{\label{cifar10}Performance of the DDDM under three attacks on the CIFAR10 dataset. 
    The same as in~\Cref{mnist}, we include the performance under individual attacks as well as the 
    overall performance with the global optimal dropout rate pair
    $(a^*,b^*)=(0.6,0.8)$.
    }
\end{table}

\subsubsection{CIFAR10 Dataset}
We further carry out experiments on the CIFAR10~\cite{krizhevsky_learning_2009} dataset to challenge our approach under a more realistic scenario. To implement the test-phase dropout classifier, we use the VGG16 architecture~\cite{simonyanVeryDeepConvolutional2015} without batch normalization. A dropout layer is added to each of the last six convolutional layers. We will discuss more the method of adding dropout layers in the supplementary material.

The results are summarized  in~\Cref{cifar10}, which includes three effective attacks in our test: PGD, L2DF, and Spatial. Note that these attacks cover both white-box and black-box attack categories. Again, DDDM effectively defends all the attacks with only a minor sacrifice on the clean accuracy. 
Compared to the MNIST case, while $H_{a^*,b^*}$ keeps outperforming $h_{a^*,b^*}$, their performance gaps become much smaller.
This may be attributed to the fact that under a large model such as VGG16, six dropout layers do not introduce sufficient randomness in the dropout classifier. Hence, DDM does not work at its full power.

\subsection{Audio Classification}

\begin{table}[!t]
    \centering
    \setlength{\tabcolsep}{1.4 mm}
    \begin{tabular}{ccc}
        \toprule
                                & clean           & imperceptible    \\\midrule
        $h_{0,0}$ (\%)          & 86.73           & 5.18             \\\midrule
        $h_{a^*,b^*}$ (\%)      & 82.91           & 68.36            \\\midrule
        $H_{a^*,b^*}$ (\%)      & {85.02}  & {70.62}   \\
        \bottomrule
    \end{tabular}
    \caption{\label{speech}Performance of the DDDM. 
    Similar to~\Cref{mnist} but for the imperceptible attack on the SpeechCommands dataset. 
    Since there is only one attack, we only includes the overall performance with 
    the global optimal dropout rate pair $(a^*,b^*)=(0.4,0.4)$.
    }
\end{table}

To evaluate the cross-domain performance of DDDM, we consider the audio classification task on SpeechCommands~\cite{wardenSpeechCommandsDataset2018} dataset, which contains 35 keywords and 105829 audio clips of those keywords. 
In the audio domain, present works focus on the classification and recognition of audio clips, which serve as the base of various applications such as voice input and voice assistants.
The existing audio adversarial attacks~\cite{carliniAudioAdversarialExamples2018,qinImperceptibleRobustTargeted2019} severely endangers the safety of such automatic speech recognition (ASR) systems. 

In our test, we adopt a simplified version of the DeepSpeech2 model~\cite{amodeiDeepSpeechEndtoEnd2015} against the attack proposed by~\cite{qinImperceptibleRobustTargeted2019}, which adds human-imperceptible perturbations to the audio waveform to mislead the model predictions.
Our DeepSpeech2 model includes a Mel-spectrogram conversion layer, followed by one 1D convolutional layer, two LSTMs, and two fully-connected layers. Dropout is only applied once after the convolutional layer. 
After training, we attacked the model with the adversarial perturbation bound of 0.05.

Results in~\Cref{speech} show that 
DDDM providea defense against the imperceptible attack with only a minor decrease in clean accuracy. Comparing the dropout classifier $h_{a^*,b^*}$ with the DDDM $H_{a^*,b^*}$, the latter show better performance under clean and adversarial case situations. We notice that the gap between $h_{a^*,b^*}$ and $H_{a^*,b^*}$ is small. This is similar to the case for CIFAR10 and could be attributed to the insufficient number of dropout layers.

\subsection{Text Classification}

\begin{table}[!t]
    \centering
    \setlength{\tabcolsep}{1.4 mm}
    \begin{tabular}{ccc}
        \toprule
                           & clean           & TextBugger       \\\midrule
        $h_{0,0}$ (\%)     & 88.44           & 1.7              \\\midrule
        $h_{a^*,b^*}$ (\%) & 89.16           & 22.10            \\\midrule
        $H_{a^*,b^*}$ (\%) & {89.41}  & {86.75}   \\
        \bottomrule
    \end{tabular}
    \caption{\label{imdb}Performance of the DDDM. 
    Similar to~\Cref{speech} but for the TextBugger attack on the IMDB dataset. 
    The global optimal dropout rate pair is $(a^*,b^*)=(0.4,0.4)$.
    }
\end{table}

At last, we evaluate our approach on the text classification task.
Natural languages consist of discrete characters and words. The process of mapping tokens to embeddings prevents directly applying attack methods from the image domain on textual classifiers. 
A text-specific attack methods have been developed
~\cite{liTextBuggerGeneratingAdversarial2019}.

The DDDM is evaluated on the IMDB dataset~\cite{maasLearningWordVectors2011} against the TextBugger attack~\cite{liTextBuggerGeneratingAdversarial2019}.
We use an LSTM 
classifier resembling the one in the TextBugger work, with adding an additional dropout layer to the pre-activation outputs of each gate in the network.
Words from the IMDB dataset are first converted into pre-trained GloVe embedding vectors of dimension $300$, then fed to the LSTM classifier.

We summarize the experimental results in~\Cref{imdb}. DDDM $H_{a^*,b^*}$ successfully protected the classifier with a considerable improvement in performance compared to the dropout classifier $h_{a^*,b^*}$.
The robust accuracy of $H_{a^*,b^*}$ under the TextBugger attack approaches its clean accuracy.
Moreover, we note that the clean accuracy of $H_{a^*,b^*}$ is higher than the one of the undefended model $h_{0,0}$.
This could be explained by the fact that we use the clean training data to evaluate the likelihood.
We emphasize that this does not harm the effectiveness of our approach since it does not require information about the possible attacks beforehand.
\section{Discussion}

The current work is merely proof of the principle that combining an evidence accumulation mechanism with deep neural networks enhances robustness and provides defense against adversarial attacks. The key to the framework's success lies in the interplay between two components of the framework. The first is the mechanism to introduce noise into the system. 
Partly because the random noise itself degrades the performance; therefore, achieving a balance between performance and robustness poses a challenge. 
The second component, the evidence accumulation in the DDM, provides a mechanism to solve this problem and allows more flexibility in terms of the level of additional noise added into the system. In addition to artificially introduced noise, noise exists in many real-world situations, such as the automatic driving systems.
In these cases, noise naturally presented in the input may be combined with artificially introduced noise for defense against attacks.

The attractiveness of the framework also lies in its simplicity. DDDM does not require the networks to be trained differently for any particular attacks. Instead, it is agnostic to the type of attack and its particular scheme of noise. Thus, it provides protection not only for the attacks tested here but also for other attacks. The framework also does not depend on particular types of networks. It can also be used in combination with more sophisticated networks and larger datasets.

Other than what has been mentioned above, several other directions to extend the current framework can be explored in the future. In addition to the dropouts, other ways of introducing noise into the system may help to battle perturbations. A mechanism of online adjustment of decision thresholds may help improve the speed-accuracy tradeoff in a dynamic environment. Finally, the evidence accumulation mechanism is not implemented with a neural network in the current study. However, previous studies have shown that simple recurrent networks may achieve the functionality~\cite{wangProbabilisticDecisionMaking2002,wongRecurrentNetworkMechanism2006}, providing an end-to-end neural network solution for the DDDM.
\section{Conclustion}

Inspired by the neural mechanism of decision making, we develop a novel framework (see~\Cref{fig1}) and demonstrate its effectiveness in improving the robustness of deep learning networks against adversarial attacks in different domains with a brain-like mechanism (see~\Cref{mnist,cifar10,speech,imdb}). In contrast to adversarial training, the DDDM shows an advantage of sustaining the same level of clean accuracy after being attacked. By extending the one-shot classification into a temporal decision-making process based on evidence accumulation, our framework trades inference time for accuracy to achieve robustness. 

To the best of our knowledge, this is the first framework to combine deep neural networks and a brain-inspired decision-making mechanism for defense against adversarial attacks and the first framework to provide cross-domain protection without adversarial training. We hope that our work brings the advances of neuroscience research to the attention of the artificial intelligence community and encourages researchers to take a closer look at the most powerful yet robust classifier —— our brain.

\section*{Acknowledgments}

This work was supported by grants from the National Science and Technology Innovation 2030 Project of China to Yi Zhou (2021ZD0202600) and to Tianming Yang (2021ZD0203701).
This work was also supported by grants from the Chinese Academy of Sciences (XDB32070100), and the Shanghai Municipal Science and Technology (2018SHZDZX05) to Tianming Yang.

\bibliography{DDDM}
\bibliographystyle{icml2021}

\appendix
\onecolumn

\section{Likelihood Estimation}

In the DDDM, model decisions are made when $\text{Pr}(H_i | {T_1,\dots,T_t })$
reached a threshold $A$. 
According to the Bayes theorem, it can be updated with likelihood $\text{Pr}({T_t }|H_i )$. 
However, this likelihood is difficult to compute directly because the probability density is continuous in $T_t$. 
Instead, we estimated the likelihood $\text{Pr}({T_t }|H_i )$ with $\text{Pr}({I_t }|H_i )$, where $I$ is the sorted combination of of the $k$ most confident labels in prediction. 
For example, in an MNIST experiment, where $T_t=(0,0,0.1,0.3,0.4,0.15,0.05,0,0,0)$ and $k$ is set to $3$. The $k$ largest predicted labels are "4" ($p=0.4$), "3" ($p=0.3$) and "5" ($p=0.15$), then $I_t$ will be "435". 
When $H_i$ is the hypothesis that "input image belongs to label $i$", $\text{Pr}({I_t }|H_i )$ refers to the frequency of $I_t$'s appearance when the neural network is fed with images of label $i$ from the training set. 
We choose $k=3$ for all experiments in this paper, except for the IMDB dataset, where $k=2$.

\section{Attack Details}

\subsection{MNIST Dataset}

We adopted the simple CNN in~\cite{madryDeepLearningModels2019} with an additional dropout layer that follows each layer except the input/output ones.
We evaluated DDDM on this network with the following attack methods:
\begin{itemize}
    \item the Fast Gradient Sign Method (FGSM) attack~\cite{goodfellowExplainingHarnessingAdversarial2015a}, with parameters $\epsilon=0.3$.
    \item the Projected Gradient Descent (PGD) attack~\cite{madryDeepLearningModels2019}, with parameters  $\epsilon=0.3$ and steps = 40.
    \item the $L_2$ Carlini and Wagner ($L_2$  C\&W) attack~\cite{carliniEvaluatingRobustnessNeural2017} with $\text{steps}_{max}=1000$.
    \item the $L_2$ DeepFool attack~\cite{moosavi-dezfooliDeepFoolSimpleAccurate2016a}, with parameters  $\epsilon=1.5$ and steps = 50.
    \item the Salt and Pepper (Salt\&Pepper) attack, with parameters  $\epsilon=10$ and steps = 1000.
    \item the repeated Additive Uniform Noise (uniform) Attack~\cite{rauberFastDifferentiableClippingAware2020}, with with parameters  $\epsilon=0.3$ and repeats = 100.
    \item the Spatial attack~\cite{engstromExploringLandscapeSpatial2019a}, with parameters $\text{translation}_\text{max}=3, \text{translation}_\text{num}=5, \text{rotation}_\text{max}=30^\circ, \text{rotation}_\text{num}=5$, and $\text{random}_{steps}=100$.
    \item the Square attack~\cite{andriushchenkoSquareAttackQueryEfficient2020}, with parameters  $\epsilon=0.3$ and $\text{steps}_{max}=100$. 
\end{itemize}

\subsection{CIFAR10 Dataset}

We use the VGG16 architecture~\cite{simonyanVeryDeepConvolutional2015} without batch normalization. A dropout layer is added to each of the last six convolutional layers.
We evaluated DDDM on this network with the following attack methods:
\begin{itemize}
    \item the PGD attack~\cite{madryDeepLearningModels2019}, with parameters $\epsilon=\frac{8}{255}\approx 0.031$ and steps=200.
    \item the $L_2$ DeepFool attack~\cite{moosavi-dezfooliDeepFoolSimpleAccurate2016a}, with parameters $\epsilon=0.12$ and steps = 50.
    \item the Spatial attack~\cite{engstromExploringLandscapeSpatial2019a}, with parameters $\text{translation}_\text{max}=3, \text{translation}_\text{num}=5, \text{rotation}_\text{max}=30^\circ, \text{rotation}_\text{num}=5$, and $\text{random}_{steps}=100$.
\end{itemize}

\subsection{SpeechCommands Dataset}

In this task we adopted a mini version of DeepSpeech2 {amodeiDeepSpeechEndtoEnd2015} model and the imperceptible ASR attack method~\cite{qinImperceptibleRobustTargeted2019}. Our DeepSpeech2 model includes a Mel-spectrogram conversion layer, followed by one 1D convolutional layer, two LSTMs and two fully-connected layers. Dropout is only applied once after the convolutional layer. The model was trained from scratch, only on the SpeechCommands dataset. After training, we attacked the model with $\epsilon=0.05$ and random target words.

\subsection{IMDB Dataset}

In this task we adopted the black-box attack method and targeted LSTM classifier in~\cite{liTextBuggerGeneratingAdversarial2019}. Dropout layers were inserted before the sigmoid or tanh operation in each of the four gates in the LSTM.

\section{Method of Adding Dropout Layers to VGG16}

Due to the relatively large size of the the VGG16 network, it is infeasible to train the network if one adds a dropout layer to each of the original layers.
Instead, we want to add dropout layers only to those parts that are sensitive to adversarial attacks. 
To discover such parts, we investigate the hidden outputs of each layer in VGG16.  We compared the hidden outputs for the clean inputs to the ones for the adversarial inputs. Those adversarial inputs are generated using PGD attack with $\epsilon=0.031$ and steps=200.

\begin{figure}[!ht]
    \centering
    \includegraphics[width=0.43\textwidth]{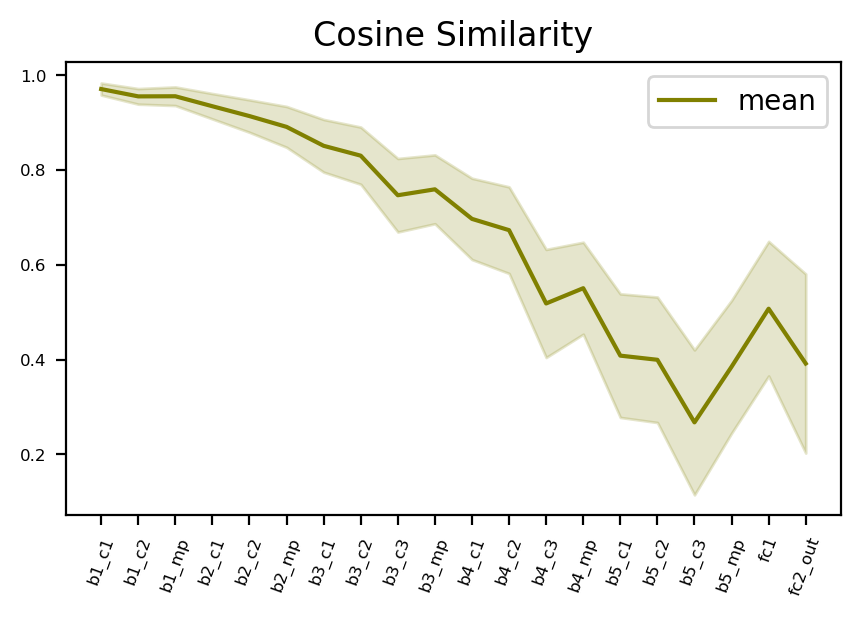}
    \includegraphics[width=0.43\textwidth]{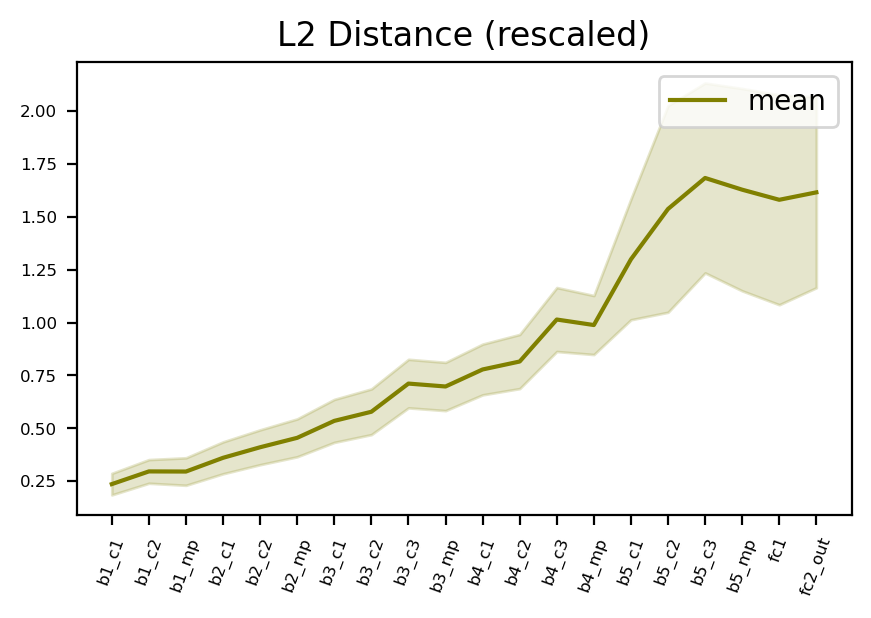}
    \caption{U
    The cosine similarity and the $L_2$ distance between hidden outputs for the clean and adversarial inputs.
    \label{dist}}
\end{figure}

\begin{figure}[!ht]
    \centering
    \includegraphics[width=0.9\textwidth]{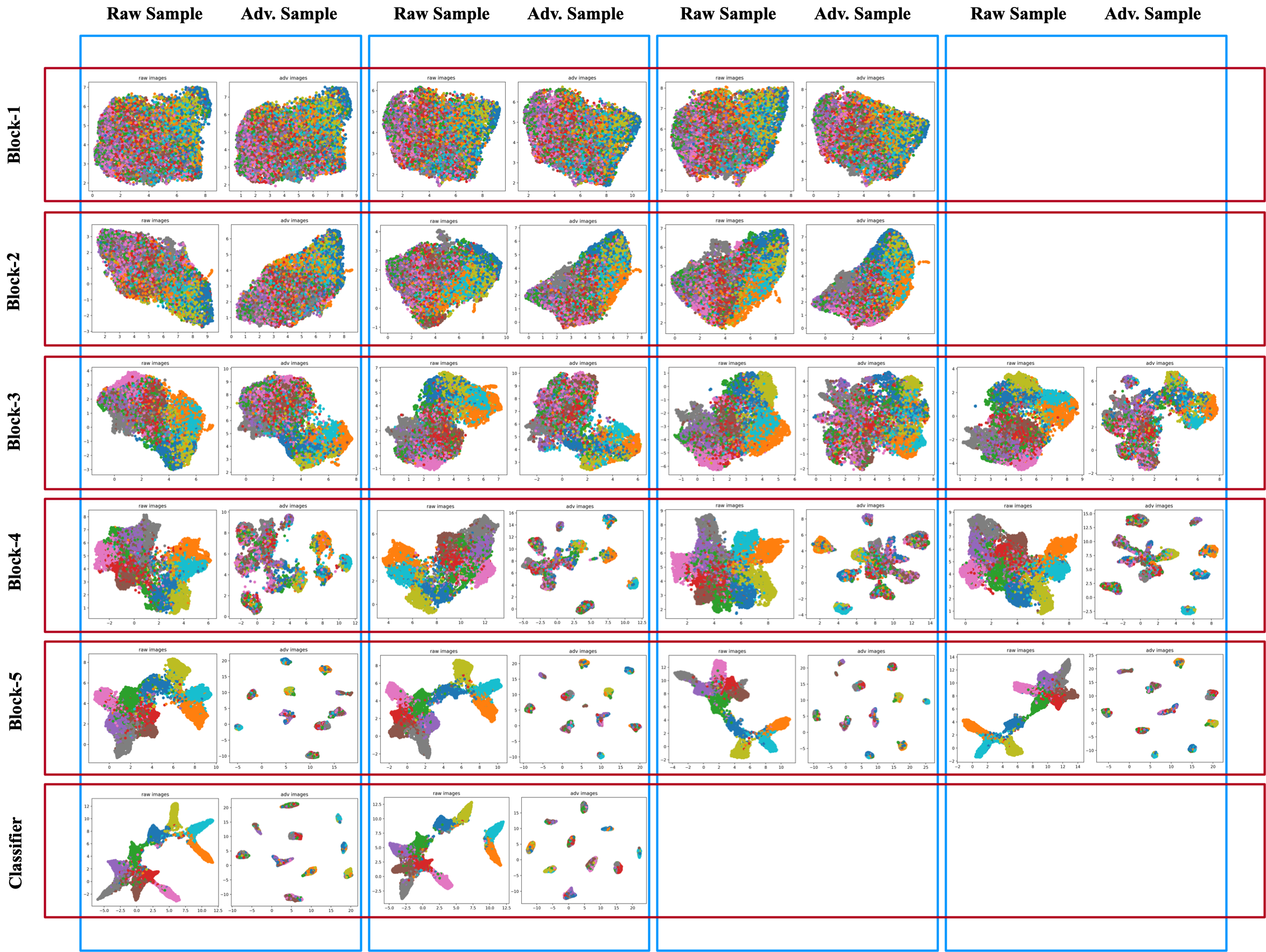}
    \caption{UMAP for hidden outputs of each layer in VGG16.\label{UMAP}}
\end{figure}

Two metrics are investigated, namely the cosine similarity and the $L_2$ distance between the hidden outputs for the clean and adversarial inputs. 
For the $L_2$ distance, we rescale the results by deviding the 
We plot the mean and standard deviation of the two metrics for each layer in~\Cref{dist}
In addition, we also plot the UMAP for the hidden outputs of each layer. The results are summarized in~\Cref{UMAP}

Both results reveal that the lower part of the network, i.e. the latter six convolutional layers are more sensitive to the adversarial inputs. Hence, we decide to add dropout layers only to those six layers.

\end{document}